# Discovering distinctive elements of biomedical datasets for high-performance exploration

Md Tauhidul Islam[1] and Lei Xing[1,*]

[1]Department of Radiation Oncology, Stanford University, Stanford, California-94305, USA

## ABSTRACT

The human brain represents an object by small elements and distinguishes two objects based on the difference in elements. Discovering the distinctive elements of high-dimensional datasets is therefore critical in numerous perception-driven biomedical and clinical studies. However, currently there is no available method for reliable extraction of distinctive elements of high-dimensional biomedical and clinical datasets. Here we present an unsupervised deep learning technique namely distinctive element analysis (DEA), which extracts the distinctive data elements using high-dimensional correlative information of the datasets. DEA at first computes a large number of distinctive parts of the data, then filters and condenses the parts into DEA elements by employing a unique kernel-driven triple-optimization network. DEA has been found to improve the accuracy by up to 45% in comparison to the traditional techniques in applications such as disease detection from medical images, gene ranking and cell recognition from single cell RNA sequence (scRNA-seq) datasets. Moreover, DEA allows user-guided manipulation of the intermediate calculation process and thus offers intermediate results with better interpretability.

## Introduction

A long-prevailed scientific question is "how do human beings recognize things?" A popular theory in attempting to answer the question is the recognition-by-components[1], which states that the visual information gained from an object is divided into simple geometric components or parts, such as blocks and cylinders, also known as "geons" (geometric ions). In matching two objects, the human brain first finds the difference between their parts. If the difference is within a threshold, the two objects are considered to be the same, and vice versa. In this manner, the human brain can recognize the same object from different viewpoints and angles. It is interesting to point out that, when multiple objects are involved, in order to find the object closer to a reference one (saved in memory), it is not necessary for human brain to find all the object parts. Instead, only differentiating object parts are sufficient for the brain to reach a decision. The idea of learning the distinct elements of data is important for recognition, classification and detection tasks in machine learning from this perspective.

Analytical techniques that learn the data/object parts or holistic representation of data/objects, such as PCA[2], NNMF[3] and variants of NNMF (convex- and semi-NNMF[4], nonsmooth NNMF[5] and graph regularized NNMF[6]), have been developed for face recognition, object detection, text similarity analysis, image ranking, document clustering and other related applications. These methods, however, suffer from undifferentiated learning of all the data elements, resulting in learning of unnecessary information and wrong detection of the data classes. On a fundamental level, the learning process of these methods is in contrast with the human brain recognition process, where only the distinct elements of data are used.

Machine learning techniques such as capsule deep networks (CapsNet) are recently proposed for similar purpose: learning the object parts. Although they overcome one of the most serious limitations of the convolutional neural network (CNN): invariance of the network towards the position of different object parts, CapsNet also has some limitations that hinder its application in high-dimensional data analysis. First, CapsNet is not able to learn complex object parts and thus is not suitable for analyzing biomedical datasets such as medical images, where the organ parts are in general very complex. Till date, CapsNet has been shown to be effective in learning object parts from very simple datasets such as hand written digits of MNIST. In complex datasets such as CIFAR-10 object database, the developed CapsNets' performance is very poor as reported in several recent works (less than 20% accuracy)[7]. Moreover, although CapsNet learns the object parts and encode the parts into lower-dimensional representation, it is not possible to interpret the lower-dimensional representation from CapsNet. As a result, the final outcome of the CapsNet is also uninterpretable. Thus CapsNet may not be suitable in biomedical and clinical applications, where interpretability is urgently needed.

There are other unsupervised deep learning approaches for learning the data representation: autoencoder[8], generative models[9] and contrastive learning approaches. Contrastive methods learn data representations using objective functions similar to those used for supervised learning, but train networks to perform pretext tasks where both the inputs and labels are derived from an unlabeled dataset[10,11]. SIMCLR (simple framework for contrastive learning of visual representations)[12] is a recent method of this class, which learns the low-dimensional representation of high-dimensional data using unsupervised instant discriminant analysis and shows significant promise in better learning the data. However, all these deep learning methods along with SIMCLR are time-consuming, requires large amount of data and most importantly not interpretable similarly as the supervised deep learning methods. Thus, these methods are not particularly suitable for biomedical data analysis, where data is scarce and decision-makers are skeptical of usage of a non-uninterpretable method[13]. Among the supervised analytical techniques, LDA[14] is the major one and has been used for detection and classification tasks for many decades. Although LDA is interpretable, it has the generic limitations of the supervised techniques: requirement of labels and large datasets. GoogLeNet[15], DenseNet[16], VGGNet[17], AlexNet[18], DarkNet[19] and NASNet[20] are the supervised deep learning networks that are being widely used for classification and recognition tasks in a wide range of scientific applications.

Here, we propose distinctive element analysis (DEA), which learns only the differentiating elements of data using contrast of correlation distance between data points in high dimension. The method possesses two salient features: 1) its learning is performed in higher dimension than the original data dimension, which ensures that no information gets lost in the learning process; and 2) it provides information about the level of distinction between two data elements. Computationally, DEA is built on a kernel-driven triple optimization network. DEA optimizes the first objective function to find a number of data clusters with centroids that minimizes the sum of correlation distances from the centroids to the associated data points in the clusters (Fig. 1 a-step 1). In general, the number of clusters is higher than the original data dimension to prevent any information loss. Each of the high-dimensional centroid points represents a specific element of data (i.e., part of an object). Correlation distances from these points separate out the same and different data elements. If the correlation distance from a cluster center to a data point is low, the data point is assumed to be from the data element the cluster represents. The correlation distance of data points from the cluster centers provides a high dimensional distance matrix. The centroids are then ranked based on a kernel optimization and centroids with high scores are retained for the next step (Fig. 1 a-step 2). This step removes the unimportant centroids from the distance matrix which helps efficient data condensation in the next step (supplementary section 3). Finally, a deep learning encoder-decoder optimization is used to compact the distance matrix with selected centroids and reduce its dimension to the final desired dimension (Fig. 1-step 3) to obtain the DEA components.

## Results

In the following, we explore the broad applications of DEA in diverse biomedical fields such medical imaging and genomic science for applications such as disease detection and cell classification. We demonstrate the efficacy of DEA in learning the differentiating elements of datasets and usage of it in the above-mentioned tasks. We will also explore how the intermediate results from DEA can be interpreted and manipulated to obtain better accuracy. We will use several established datasets from different biomedical disciplines for these purpose, i.e., diabetic retinopathy and chest x-ray datasets and single cell RNA-seq data acquired from retinal bipolar cells. In the end, we will analyze the computational perspective of DEA in comparison to other traditional techniques and discuss the limitations of DEA.

### DEA performs accurate classification of diabetic retinopathy images

We choose a diabetic retinopathy dataset as the first example for showing efficient classification task by DEA. This dataset is a collection of high-resolution retina images taken under a variety of imaging conditions. Left and right fields' images are provided for every subject. A clinician has rated the presence of diabetic retinopathy (DR) in each image on a scale of 0 to 4, according to the following scale: 1) 0 - No DR, 2) 1 - Mild, 3) 2 - Moderate, 4) 3 - Severe and 5) 4 - Proliferative DR. The images in the dataset come from different models and types of cameras, which may have affected the visual appearance of left vs. right. Some images are shown as one would see the retina anatomically (macula on the left, optic nerve on the right for the right eye). Others are shown as one would see through a microscope condensing lens (i.e. inverted, as one sees in a typical live eye exam). The number of images used in our analysis is 10,538 (randomly selected 30% of the total images).

The PCA, NNMF, and DEA components are shown in Fig. 2 (a-d) for the diabetic retinopathy dataset along with some randomly chosen images. It is seen that PCA computes the holistic representation of the images, whereas NNMF finds out the different important parts of retina in the components. DEA computes the differentiating elements of data between classes. In many components from DEA, it is seen that the background parts are highlighted, which can be easily removed to further improve the accuracy. The classification accuracy for different methods is presented in (e1-e3) for component number of 8, 16, and 32. It is seen from these figures that DEA performs best and much better than PCA and NNMF. As an example, for 8 component classification, DEA provides accuracy improvement of more than 40% when only 30% data is used for training. The accuracy for CapsNet, GoogleNet and DenseNet for 70% training data is 67%, 73% and 79%, respectively. Thus, DEA shows higher accuracy than CapsNet and GoogLeNet and comparative accuracy to DenseNet.

### DEA offers accurate classification of chest x-ray images

As the next example, we choose the NIH-released ChestX-ray14 dataset[21], which contains 112,120 frontal-view chest X-ray images of 30,805 unique patients. Wang et al.[21] annotate each image with up to 14 different thoracic pathology labels using automatic extraction methods on radiology reports: Atelectasis, Cardiomegaly, Effusion, Infiltration, Mass, Nodule, Pneumonia, Pneumothorax, Consolidation, Edema, Emphysema, Fibrosis, Pleural, Thickening and Hernia. We randomly choose 10% (11,211) of the x-ray images to analyze by DEA. Before analyzing the images, we downscale the images to $224 \times 224$ and normalize based on the mean and standard deviation.

The PCA, NNMF, and DEA components are shown in Fig. 3 (a-d) for the chest X-ray dataset along some randomly chosen images. Similar to diabetic retinopathy dataset, PCA computes a overall representation of the X-ray images, whereas NNMF finds out the parts of the images. DEA on the other hand finds out the parts of images, where the distinction of data classes are present. The classification accuracy is shown in (e1-e3), where it is seen that DEA performs superiorly in comparison to all other methods. Specially for component number of 32, we see that DEA provides an accuracy improvement of at least 20% in comparison to all the competing methods. This proves the efficiency of DEA in finding the image elements that helps best in correct classification. The accuracy for CapsNet, GoogLeNet and DenseNet for 70% training data is 46%, 51% and 54%, respectively for chest x-ray dataset. Thus, DEA shows higher accuracy than CapsNet and GoogLeNet and similar accuracy to DenseNet.

In summary, DEA performs much better in classification of medical images in comparison to PCA, NNMF and CapsNet in all cases and better or comparable to supervised techniques GoogLeNet and DenseNet.

**DEA finds out most important genes and offers highly accurate cell classification from scRNA-seq data**

The dataset analyzed in this section is from a massively parallel scRNA-seq profiled from a heterogeneous class of neurons, mouse retinal bipolar cells (BCs)[22]. The main motivation of the study is that neuronal types can be characterized and classified using gene expression data acquired from the neuron cells. From a population of 27499 BCs, the authors derived a molecular classification that identified 15 types, including all types observed previously and two new types. The experimental protocol followed by the authors is as follows: 1) retinas from Vsx2-GFP[22] mice were dissociated, followed by fluorescence activated cell (FAC) sorting for GFP+cells. 2) Single-cell libraries were prepared using Drop-seq and sequenced. 3) Raw reads were processed to obtain the digital expression matrix (genes $\times$ cells). The scRNA-seq data is pre-processed and the 500 most variable genes are retained for further analysis by PCA, NNMF, and DEA.

The importance score of the 500 genes estimated using different methods are shown Fig. 4 (a). The three rows of this figure represents the results from PCA, NNMF, and DEA, respectively. It is seen from these figures that a there are some common genes selected by by all three methods as important. The number of most important genes (have score of more than 0.5 (mean value)) is less in case of DEA results than others. Moreover, the difference between importance scores of important and non-important genes is higher in case of DEA results than other methods' results, which makes it easier to find out the important genes using DEA.

To measure the performance of different methods in finding the most important genes, we selected the first 30 and 50 genes with highest scores and use just those genes in analysis of the scRNA-seq data by t-SNE and results are shown in Fig. 4 (b1-d2). The visualization of the original data is shown in (a1), where it is seen that there are two clusters in the results and most of the data classes are completely inseparable. This is because of the unnecessary and conflicting information from the unimportant genes and the dropout effect from them. However, when we just selected 30 most important genes using DEA and analyzed the data, we see that all the classes get separated Fig. 4 (b1). For the first 50 genes, the results get better as it includes more information for separation of the data classes Fig. 4 (c1). The results from PCA and NNMF are shown in Fig. 4 (a2,c2), (b2,d2), respectively. However, the data classes are not separable in these visualizations as it is in DEA results. Therefore, DEA performs much better in selecting the most important genes in comparison to other competing methods.

Next, we analyze the projected original data onto the components from DEA, NNMF, and PCA using t-SNE in Fig. 5. The visualizations are presented in (a1-c1) for PCA, NNMF, and DEA for 8 reduced components, respectively. Even for this low number of components, it is seen that the DEA results clearly separate out the data classes, which results in high accuracy in classification tasks. As the component number increases, results of all methods improved (a2-c3). However, for every cases, DEA performs much better than PCA and NNMF. This observation is reflected on the quantitative performance of different methods shown in Fig. 5 (a4-c4).

In summary, DEA successfully finds out the most important genes from scRNA-seq data and computes components that allow highly accurate classification.

## Discussion

High-dimensional data possess inherent patterns with similar and different elements when compared between data classes[23]. Discovering these data elements are fundamental for finding the patterns and explaining them intuitively. Moreover, these similar and distinctive data elements are useful for reaching a decision from the data or becoming confident about the decision made by a machine or algorithm. In this work, a DEA framework is established to discover the distinctive elements of high-dimensional datasets and its superior performance is demonstrated in different perception-driven tasks such as image classification, gene ranking and cell recognition.

The use of deep learning techniques in high-dimensional data analysis is limited because of two inherent limitations[13,24]. Firstly, deep learning techniques produce non-interpretable results and does not offer any explanation of final or intermediate results[25,26]. Secondly, deep learning techniques are mostly supervised and require large amount of labeled data, which is expensive in many cases and are not available in many scenarios[27–32]. Although powered by deep learning, DEA offers interpretable results and requires less data because of its unsupervised nature. The intermediate results from DEA are interpretable and clearly show the differentiating parts of objects or datasets. DEA also offers the user to fully control which components to use in the downstream tasks. Therefore, DEA is a better choice in many cases than the deep learning methods such as SIMCLR[12], CapsNet, GoogLeNet and Densenet, which produce uninterpretable results.

In recognition tasks, one of the main concerns is the background noise in the datasets[33,34]. In image based recognition or classification, the background noise comes as different objects which are unnecessary and creates serious obstacles to correctly recognize the actual object. DEA finds out the background objects in different components, which provides an unique opportunity to remove them from downstream tasks. As we demonstrated in Fig. S6 in case of Yale face database, diabetic retinopathy and chest X-ray datasets, such removal of noisy backgrounds result in improvement of classification performance.

Single cell RNA sequence data is inherently noisy because of the dropout effect and as a result selection of highly variable gene and cell recognition become very challenging[35–37]. DEA is particularly suitable for this task because of its inherent denoising characteristics. Different to traditional techniques which work on the raw RNA-seq data, DEA works on the gene-contrast data[38]. The computation of contrast of gene expression in DEA automatically removes the noise[39–42]. Thus, DEA is able to offer better performance in selection of important genes as well as cell recognition tasks from gene expression data.

The main limitation of DEA is that its usage is limited to datasets with distinctive elements. Analysis of a single class data with DEA is meaningless as there is no contrast in the data to analyze. PCA and NNMF are better choice to analyze these datasets. Another limitation of DEA is the larger computational cost in computing the components in comparison to methods like PCA and NNMF. The higher computational cost of DEA comes from the use of deep learning optimization in the third step. However, we note that the computational cost of DEA is reasonable (in the order of minutes) and at least one-order less than the traditional deep learning techniques (such as SIMCLR, GoogLeNet and DenseNet), which have computation requirements in the order of hours/days. We report and discuss the computational cost for three different methods for datasets with different number of data points and dimensionality in the section 2 of the supplementary.

We demonstrated usage of DEA in a few popular tasks, where interpretation is as important as the performance. However, there are many other applications, where DEA can be very promising. Three such applications are face recognition[43], drug discovery[44] and activity recognition from sensor data[45–47]. We demonstrated the high performance application of DEA for face recognition in supplementary section 1. In drug discovery, finding the similar drugs in terms of effect or molecular structure is important. As DEA finds out the differentiating elements of the data, it can be used to find out the differentiating parts of two/more drugs, which might help us in finding the similar drugs with similar molecular characteristics or their effect on human body. In activity recognition tasks which is extremely important in human-model robotics[45], the DEA can be trained on some sensor data in unsupervised manner and the testing data can be classified with high accuracy. Other interesting applications of DEA can be searching similarity in images or documents, ranking data, images and documents and so on.

In DEA, we used three optimization processes to compute the accurate distinctive elements. The rationale behind selection of each of the optimization process has been discussed in the supplementary section 3 with associated results. We also used correlation distance as the measure of similarity between data elements because in most high-dimensional data disciplines, the generated data is in general correlated with each other. However, in many applications, where data elements may differ in terms of other distance metrics such as Euclidean, cosine or Minkowski. In those applications, DEA with the appropriate distance metric should be used for higher performance and correct interpretation of the results.

In conclusion, a novel DEA method for learning the distinctive elements of data has been presented to analyze the correlation between data points in high dimension. We have shown some interesting applications of DEA in medical image classification, cell recognition and visualization tasks. Because of the exceptional ability of DEA in learning the distinctive elements between data classes, the approach is capable of achieving very high accuracy in recognition tasks. Moreover, the DEA offers interpretability of its intermediate results and allows the user to manipulate the intermediate results for better performance, which is particularly attractive in biomedical and clinical applications. The DEA promises to change fundamentally the way the high-dimensional data are explored currently for classification, recognition and ranking tasks.

## Methods

### First optimization

Let us assume a data matrix X of size $M \times N$, which can be treated as M (1-by-N) row vectors $x_1, x_2, \ldots, x_M$. The data matrix is formed by taking each data point (such as an image) and put all the features of the data point (pixels for an image) in a column. Thus, the data matrix has features in rows and observations in columns. $U$ number of cluster centers in DEA[48] were computed optimizing the following function to find the $M \times U$ indicator matrix $A_{opt}$ such that:

$$A_{opt} = \arg \min_{A \in \mathscr{A}} \sum_{i=1}^{i=M} d(X_i, A_i A^T X). \tag{1}$$

The optimal value of the objective function is

$$F_{opt} = \min_{A \in \mathscr{A}} \sum_{i=1}^{i=M} d(X_i, A_i A^T X) = \sum_{i=1}^{i=M} d(X_i, A_{opt,i} A_{opt}^T X). \tag{2}$$

In the above $\mathscr{A}$ denotes the set of all $n \times k$ indicator matrices $A$. The correlation distance ($d()$) between a vector $x_s$ and the cluster center $y_t$ in the above equation can be written as

$$d(x_s, y_t) = 1 - \frac{(x_s - \bar{x}_s)(y_t - \bar{y}_t)'}{\sqrt{((x_s - \bar{x}_s)(x_s - \bar{x}_s'))}\sqrt{((y_t - \bar{y}_t)(y_t - \bar{y}_t)')}}. \tag{3}$$

The optimum distance matrix $Y \in \mathbf{R}^{U \times N} = \min_{A \in \mathscr{A}} d(X, E)$ is the result from the first optimization in DEA, which contains the distance of the data points from the centroids. Here, $E$ denotes the unique rows of $AA^T X$.

### Second optimization

**Construction of neighborhood distance matrix** The neighbor distance matrix consists of the distance value between the neighbors in a graph[49]. Let $Y$ be a graph denoting the high dimensional distance matrix from the first step of DEA, $Y = \{y_1, y_2, \ldots, y_M\}$. The neighborhood distance matrix considering $k$ number of neighbors for each data point can be written as,

$$\begin{aligned} D^k(Y) &= [d_{ij}], \text{ if } y_i \text{ and } y_j \text{ are neighbors}, (1 \leq i \leq M, 1 \leq j \leq M, i \neq j), \\ &= 0, \text{ otherwise}, \end{aligned} \tag{4}$$

where $d_{ij}$ denotes the Euclidean distance between $y_i$ and $y_j$.

*Properties of distance matrix*
A distance metric on a data matrix $\mathbf{Z}$ is a function defined as[50]

$$d: \mathbf{Z} \times \mathbf{Z} \to [0, \infty),$$

where $[0, \infty)$ is the set of non-negative real numbers and for all $\mathbf{x}, \mathbf{y}, \mathbf{z} \in \mathbf{Z}$, the following three axioms are satisfied:[50]
1. If $\mathbf{x} \neq \mathbf{y}$, then $d(\mathbf{x}, \mathbf{y}) > 0$, and $d(\mathbf{x}, \mathbf{x}) = 0$     minimality
2. $d(\mathbf{x}, \mathbf{y}) = d(\mathbf{y}, \mathbf{x})$,     symmetry
3. $d(\mathbf{x}, \mathbf{y}) \leq d(\mathbf{x}, \mathbf{z}) + d(\mathbf{z}, \mathbf{y})$,     triangle inequality

Thus, a distance matrix is always non-negative real-valued symmetric matrix.

*Formation of kernels*

**Schoenberg's theorem.** For a symmetric function $\Psi: X \times X \to \mathbb{R}$ with $\Psi(x,x) = 0$ for all $x$ the following are equivalent[51,52]:
1. $\Psi$ is a kernel of conditionally negative type.
2. The function $K(x,y) = \exp(-\frac{\Psi(x,y)}{\sigma^2})$ is a positive semidefinite kernel for all $\sigma$.

Based on this theorem and symmetry property of the distance matrix, a kernel similarity matrix is formed by computing, $S(x,y) = \exp(-\frac{D(x,y)}{\sigma^2})$.

*Optimization*

The second optimization in DEA is performed over the value of σ to maximize the following cost function:

$$\arg\max_{\sigma} \sum_{r=1}^{r=Q} s_r, \tag{5}$$

where $s_r$ is a vector of Q-maximum value of $\frac{d_r^T S d_r}{d_r^T D_g d_r}$. Here, $d_r$ is each column of $D$ and $D_g$ is an $M \times M$ diagonal matrix obtained by summing the rows of the similarity matrix $S$. $i$th diagonal element of $D_g$ is defined as $D_g(i,i) = \sum_{j=1:M} S_{i,j}$. The resulting matrix with $Q$ columns forms a distance matrix of $D_q$ with size $M \times Q$.

## Third optimization

We use the third optimization to project the high dimensional distinctive elements obtained from the second optimization to a lower dimension. To achieve this, we use encoder-decoder format[53] (Fig. 1-step 4). The encoder stage takes the input $\boldsymbol{x}_d \in \mathbb{R}^Q$ ($\boldsymbol{x}_d$ is a row of $D_q$) and maps it to a low dimensional space $\mathbf{h} \in \mathbb{R}^P$, where $U \geq Q > P$, $\mathbf{h}$ is defined as

$$\mathbf{h} = \Sigma(\mathbf{W}\mathbf{x_d} + \mathbf{b}). \tag{6}$$

Here, Σ is an activation function (such as sigmoid or linear function). $\mathbf{W}$ is a weight matrix and $\mathbf{b}$ is a bias vector. Weights and biases are usually initialized randomly, and then updated iteratively during training using backpropagation technique. After that, the decoder stage maps the low dimensional variable $\mathbf{h}$ to the reconstruction data $\mathbf{x'_d}$. This process can be written as

$$\mathbf{x'_d} = \Sigma'(\mathbf{W}'\mathbf{h} + \mathbf{b}'). \tag{7}$$

While training the encoder-decoder configuration, mean squared error between the original input and the deocder output are minimized. The squared error (also called loss) can be expressed as

$$\mathscr{L}(\mathbf{x_d}, \mathbf{x'_d}) = \|\mathbf{x_d} - \mathbf{x'_d}\|^2. \tag{8}$$

After we replace $x'_d$ in equation 8 using equation 7, we obtain

$$\mathscr{L}(\mathbf{x_d}, \mathbf{x'_d}) = \|\mathbf{x_d} - \sigma'(\mathbf{W}'\mathbf{h} + \mathbf{b}')\|^2. \tag{9}$$

If we replace $\mathbf{h}$ in equation 9 using equation 6, we obtain the following expression of the loss function:

$$\mathscr{L}(\mathbf{x_d}, \mathbf{x'_d}) = \|\mathbf{x_d} - \sigma'(\mathbf{W}'(\sigma(\mathbf{W}\mathbf{x_d} + \mathbf{b})) + \mathbf{b}')\|^2. \tag{10}$$

The resulting latent space $\mathbf{h}$ is the desired dimensionality-reduced data. To improve the performance of the this deep learning model, two additional terms ($L_2$ regularization and sparsity regularization terms) are generally added to the loss function as follows:

$$\mathscr{L}(\mathbf{x_d}, \mathbf{x'_d}) = \frac{1}{N}\|\mathbf{x_d} - \sigma'(\mathbf{W}'(\sigma(\mathbf{W}\mathbf{x_d} + \mathbf{b})) + \mathbf{b}')\|^2 + \lambda \Omega_w + \beta \Omega_s, \tag{11}$$

where $N$ is the total number of training examples and $\lambda$ and $\beta$ are coefficients of $L_2$ regularization and sparsity regularization terms. The sparsity regularization term $\Omega_s$ is defined as

$$\Omega_s = \sum_{i=1}^{P} KL(\rho\|\hat{\rho}_i) = \sum_{i=1}^{P} \rho \log(\frac{\rho}{\hat{\rho}_i}) + (1-\rho)\log(\frac{1-\rho}{1-\hat{\rho}_i}). \tag{12}$$

Here $\rho$ is the sparsity proportion parameter denoting the desired value of the average activation value of the neurons. Average activation value of $i$-th neuron is computed as

$$\hat{\rho}_i = \frac{1}{N}\sum_{j=1}^{N} \sigma(W_i^T x_j + b_i), \tag{13}$$

where $x_j$ is the $j$-th training example. In equation 11, $\Omega_w$ denotes the $L_2$ regularization term and is defined as

$$\Omega_w = \frac{1}{2}\sum_{j=1}^{N}\sum_{i=1}^{Q}(W_{ji})^2. \tag{14}$$

$\mathbf{h}$ are the rows of DEA output matrix $D_d$, which is of size $M \times P$. Here, $P$ is the desired number of DEA components.

**Implementation and parameter settings**

Matlab R2020a (Mathworks Inc., Natick, MA, USA) was used to implement the DEA technique. t-SNE implemented by Matlab have been used to produce the results of these methods.

In producing all the results in this paper, we used $U = 2N$ in DEA when $M > 2N$. When $M < 2N$, we used $U = N$. For the second optimization in DEA, the values of $\sigma$ considered were 0.5 to 1.5 in an interval of 0.1. We selected 80% of the centroids with top scores at this step of DEA, i.e., $Q = U * 0.8$. The number of neighbors ($k$) in creating the similarity matrix considered were logarithm of $M$. In case of the third optimization of DEA, the saturated linear function was used as decoder transfer function and logistic sigmoid function as encoder transfer function. The maximum number of epochs was 20,000. L2 weight regularization parameter was chosen as 0.001 and sparsity proportion parameter was set to be 0.05. The sparsity regularization parameter was chosen as 1.6.

For classification of medical images and genomic data, both the training and testing data were projected onto PCA, NNMF, and DEA components. A multiclass error-correcting output codes (ECOC) model with support vector machine (SVM) binary learners was trained in Matlab with default parameters using the projected training data. The projected testing data were then classified using the trained model.

**Data and code availability**

The datasets generated during and/or analyzed during the current study are available within the manuscript and supplementary.

## Acknowledgments

This work was partially supported by NIH (1R01 CA223667 and R01CA227713) and a Faculty Research Award from Google Inc.

## Author contributions statement

L.X. conceived the experiment(s), M.T.I conducted the experiment(s), M.T.I. analyzed the results. Both authors reviewed the manuscript.

## Competing interests

The authors declare no competing interests.

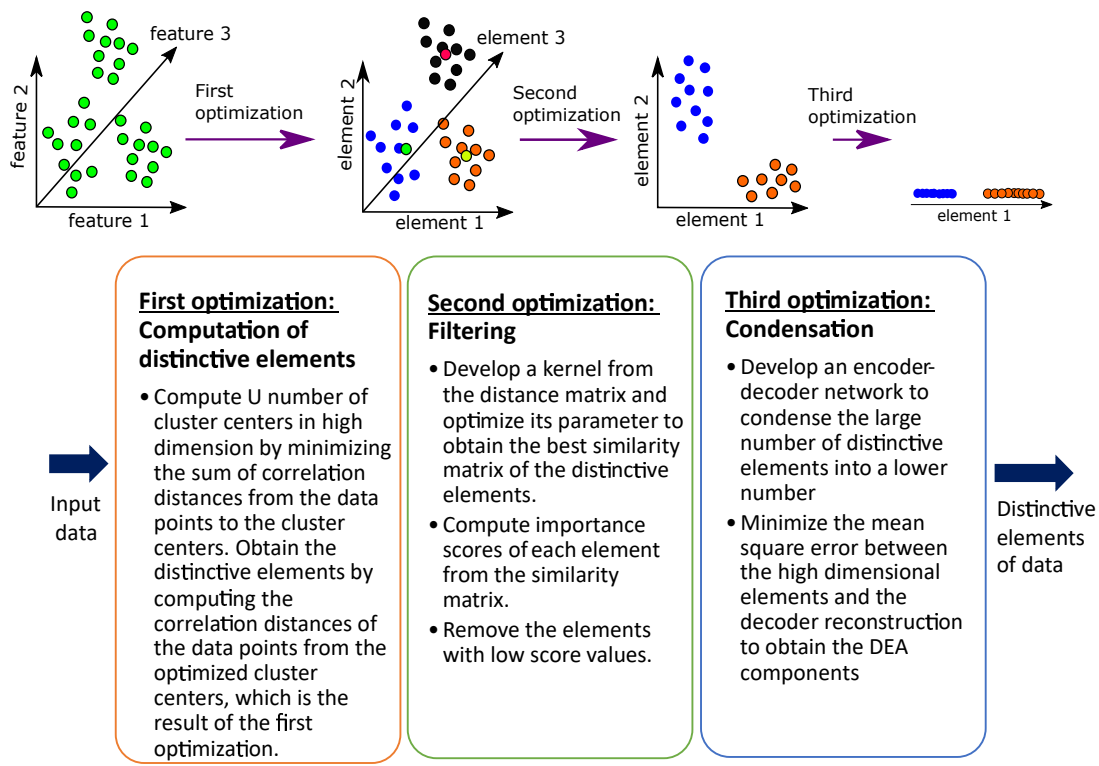

**Figure 1.** Workflow of DEA. DEA learns the distinctive elements of data based on the correlation distance between the data points and their cluster centers in high dimension using the first optimization. Each cluster center represents an element. Lower correlation distance between the data points and the cluster center implies that they are from the same data element and vice versa. DEA uses the second optimization to filter out less important distinctive elements. In the end, DEA uses the third optimization to condense the distinctive elements onto a lower dimension.

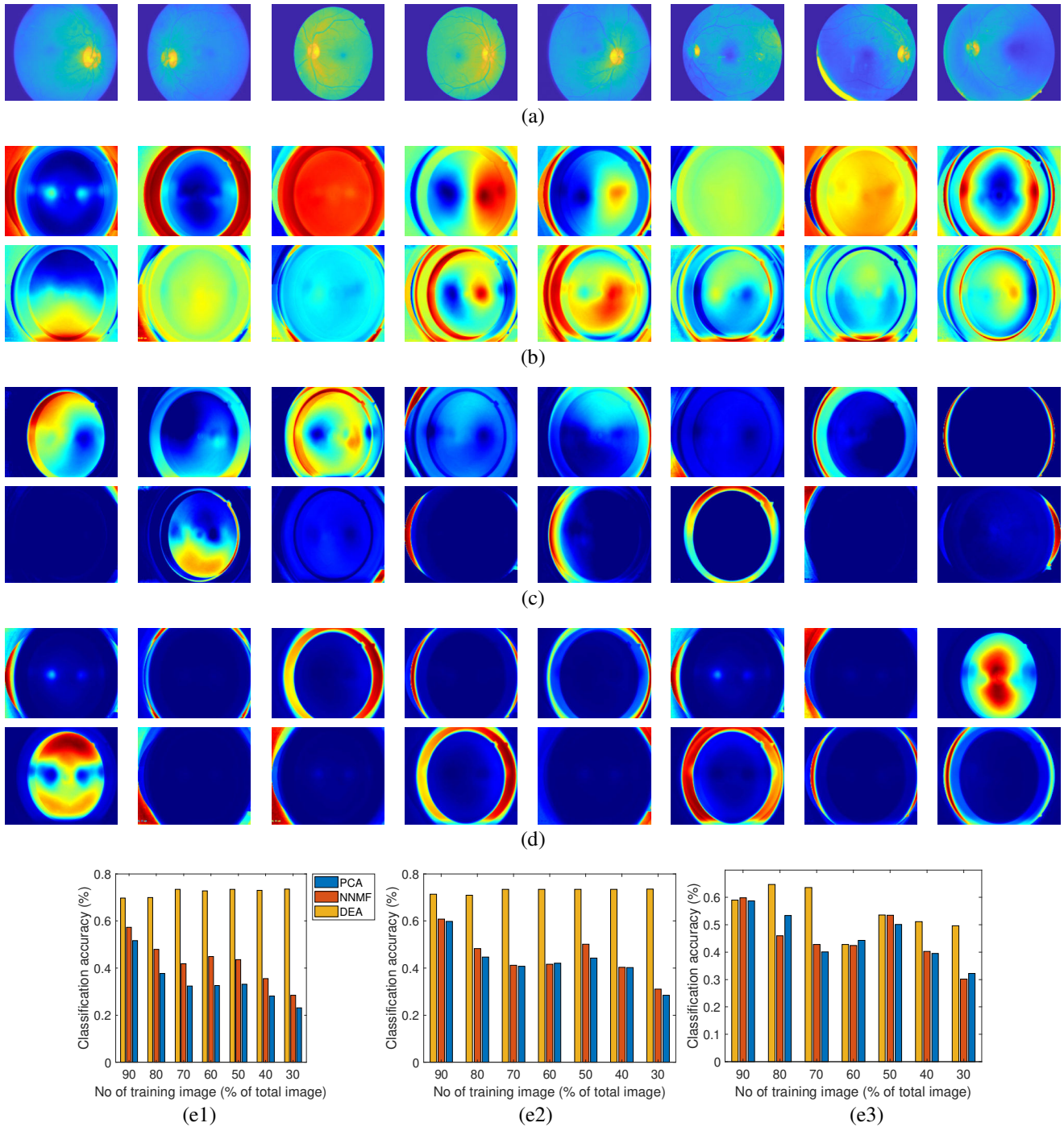

**Figure 2.** (a) 8 randomly selected images of diabetic retinopathy data (b-left to right) First 16 eigen images (principle components). (c-left to right) 16 component images from NNMF. (d-left to right) 16 component images from DEA. The number of reduced components by all the methods was 32. Holistic representation is achieved in case of eigen images. Part based representation is seen in NNMF images. In case of DEA, we see that only distinctive features of the images at different parts are seen. The level of distinction is can also be realized from DEA images. NNMF images also have different levels of intensity. However, these intensity differences do not represent any distinction levels, rather they are representatives of the image intensity at different parts and linear combination of them produces the original image. Classification accuracy by PCA, NNMF, and DEA for component numbers of 8, 16, and 32 (e1-e3). The accuracy for CapsNet, GoogleNet and DenseNet for 70% training data is 67%, 73% and 79%, respectively.

**11/14**

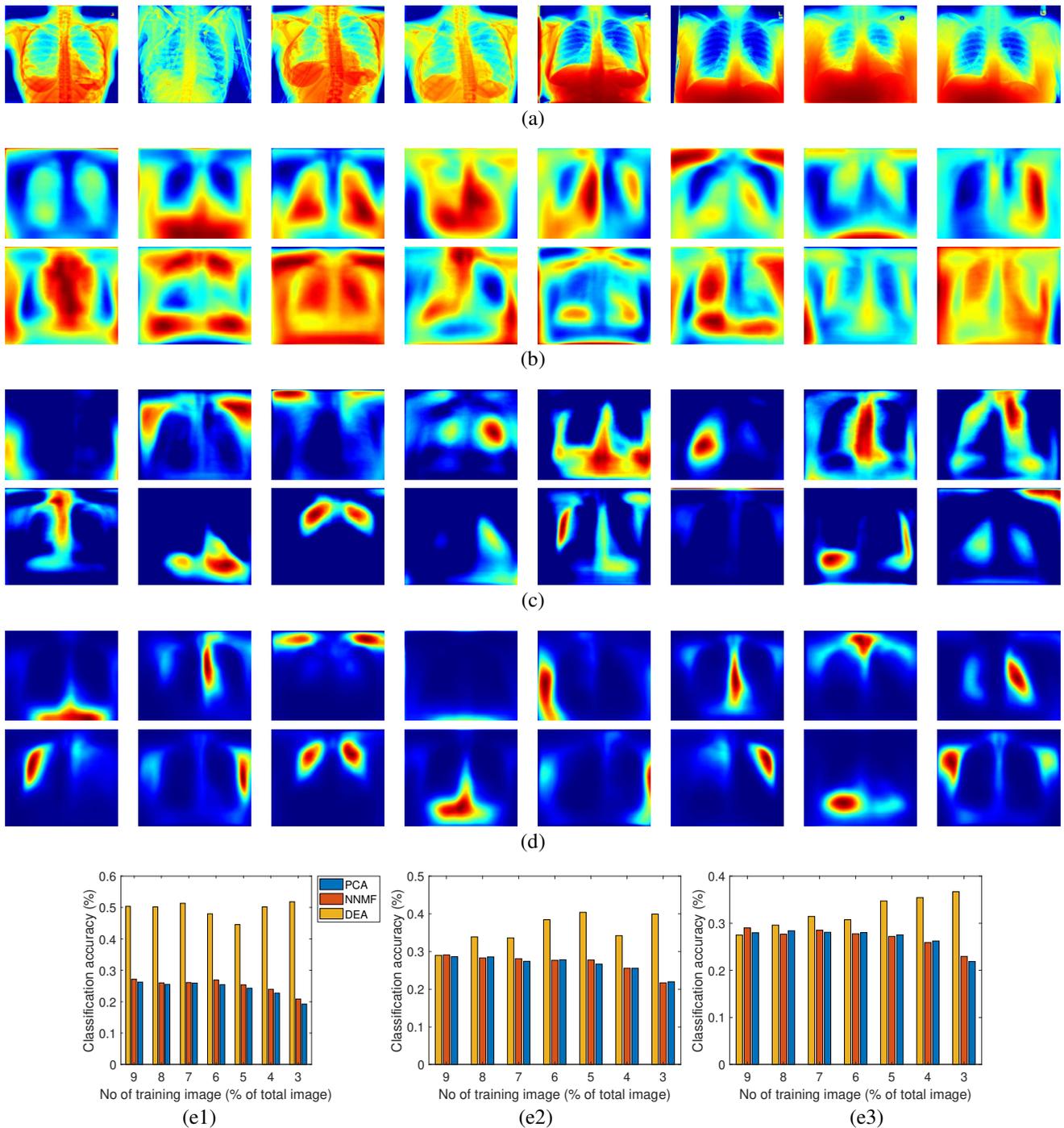

**Figure 3.** (a) 8 randomly selected images of chest x-ray (b-left to right) First 16 eigen images. (c-left to right) 16 component images from NNMF. (d-left to right) 16 component images from DEA. The number of reduced components by all the methods was 32. Holistic representation is achieved in case of eigen images. Part based representation is seen in NNMF images. In case of DEA, we see that only distinctive features of the images at different parts are seen. The level of distinction is can also be realized from DEA images. Classification accuracy by PCA, NNMF, and DEA for component numbers of 32, 64 and 96 (e1-e3). The accuracy for CapsNet, GoogleNet and DenseNet for 70% training data is 46%, 51% and 54%, respectively.

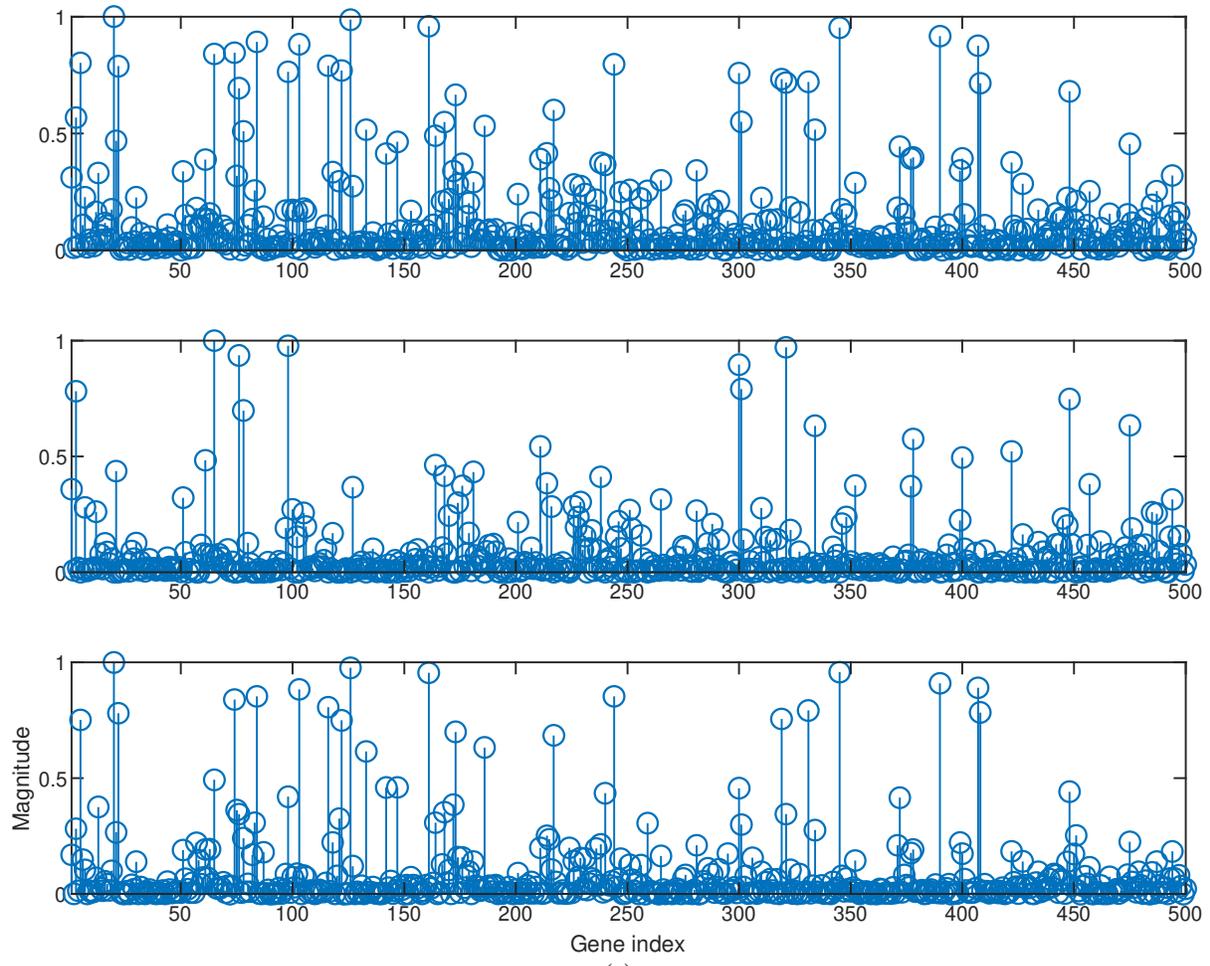

**Figure 4.** (a) Magnitude of components for each gene in scRNA-seq data from retinal bipolar neurons. PCA, NNMF, and DEA components are shown in first-third row, respectively. t-SNE visualization of the original data with 500 genes is shown in (a1). t-SNE visualization of the data with 30 and 50 genes selected by DEA is shown in (b1) and (c1). t-SNE visualization of the data with 30 and 50 genes selected by PCA is shown in (a2) and (c2). t-SNE visualization of the data with 30 and 50 genes selected by NNMF is shown in (b2) and (d2).

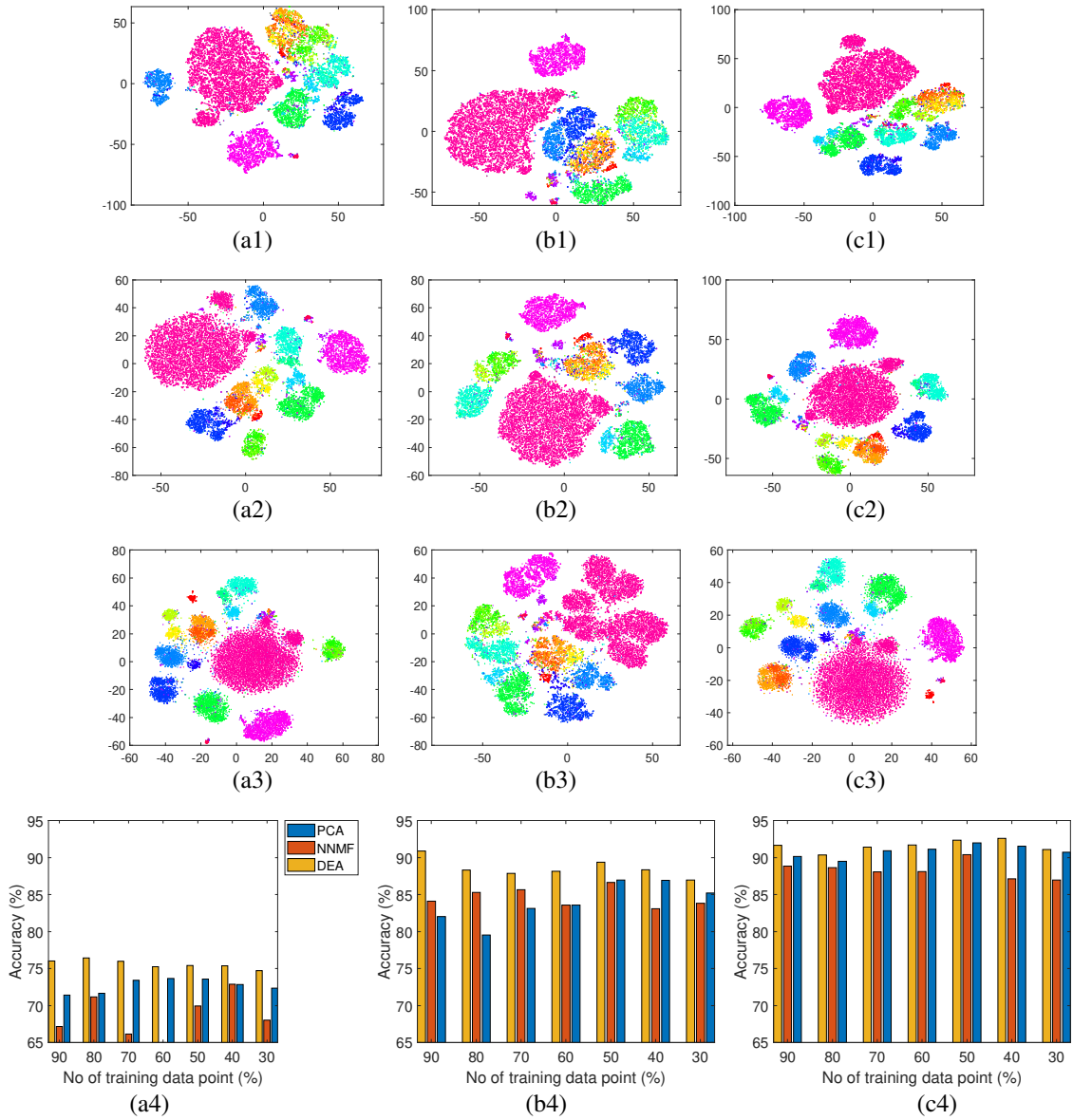

**Figure 5.** t-SNE visualizations of the projected data onto 8 (1), 12 (2), and 20 (3) components from PCA (a), NNMF (b), and DEA (c). Classification accuracy of retinal data for 4, 8, and 12 components of PCA, NNMF and DEA are shown in (a4-c4), respectively.